\documentclass[letterpaper, 10 pt, conference, twocolumn]{ieeeconf}
\IEEEoverridecommandlockouts
\usepackage{graphicx}
\usepackage{hyperref}
\usepackage[dvipsnames]{xcolor}
\usepackage{soul}
\usepackage{siunitx}
\usepackage[font=footnotesize]{caption}
\usepackage{subcaption}
\usepackage{ragged2e}

\definecolor{matlab-blue}{HTML}{0072BD}
\definecolor{matlab-orange}{HTML}{D95319}
\definecolor{matlab-green}{HTML}{77AC30}
\definecolor{faded-blue}{HTML}{C9DFFF}
\sethlcolor{faded-blue}

\hypersetup{
colorlinks=true,
citecolor=black,
linkcolor=black,
urlcolor=matlab-blue,
}

\graphicspath{{./figures/}}

\makeatletter
\def\bstctlcite{\@ifnextchar[{\@bstctlcite}{\@bstctlcite[@auxout]}}
\def\@bstctlcite[#1]#2{\@bsphack
	\@for\@citeb:=#2\do{%
		\edef\@citeb{\expandafter\@firstofone\@citeb}%
		\if@filesw\immediate\write\csname #1\endcsname{\string\citation{\@citeb}}\fi}%
	\@esphack}
\makeatother

\title{Planning and Control for a Dynamic Morphing-Wing UAV Using a Vortex Particle Model}
\author{Gino Perrotta$^{1}$, Luca Scheuer, Yocheved Kopel, Max Basescu, Adam Polevoy, Kevin Wolfe, Joseph Moore
	\thanks{*At the time of this work, all authors were with The Johns Hopkins University Applied Physics Laboratory, 11100 Johns Hopkins Road, Laurel, Maryland 20723.}%
	\thanks{$^{1}$Corresponding author. {\tt\small gino.perrotta@jhuapl.edu}}%
}

\begin{document}
\bstctlcite{IEEEexample:BSTcontrol}

\onecolumn 
\pagestyle{empty} 
\vspace*{\fill}
\begin{justify}
	\copyright 2023 IEEE.  Personal use of this material is permitted.  Permission from IEEE must be obtained for all other uses, in any current or future media, including reprinting/republishing this material for advertising or promotional purposes, creating new collective works, for resale or redistribution to servers or lists, or reuse of any copyrighted component of this work in other works.
\end{justify}
\vspace*{\fill}
\twocolumn 
\setcounter{page}{1} 

\maketitle

\begin{abstract}
Achieving precise, highly-dynamic maneuvers with Unmanned Aerial Vehicles (UAVs) is a major challenge due to the complexity of the associated aerodynamics. In particular, unsteady effects---as might be experienced in post-stall regimes or during sudden vehicle morphing---can have an adverse impact on the performance of modern flight control systems. In this paper, we present a vortex particle model and associated model-based controller capable of reasoning about the unsteady aerodynamics during aggressive maneuvers. We evaluate our approach in hardware on a morphing-wing UAV executing post-stall perching maneuvers. Our results show that the use of the unsteady aerodynamics model improves performance during both fixed-wing and dynamic-wing perching, while the use of wing-morphing planned with quasi-steady aerodynamics results in reduced performance. While the focus of this paper is a pre-computed control policy, we believe that, with sufficient computational resources, our approach could enable online planning in the future.
\end{abstract}

\section{Introduction}
The unsteady aerodynamics of post-stall flight pose a challenge for model-based control of aerobatic Unmanned Aerial Vehicles (UAVs) \cite{Sobolic2009}.
UAV dynamics models typically represent the aerodynamic state only as a wind velocity vector; acceleration of the UAV is computed from the vehicle's velocity and orientation relative to the air around it.
This approach is ideal for vehicles with high inertia flying at low incidence angles, each of which reduces the unsteady aerodynamic effects on vehicle dynamics.
With some effort, though, this approach can be extended to agile UAVs at arbitrary orientations \cite{Khan2016}.
This produces computationally efficient models, but limits expressiveness of the resulting dynamics.
Transient aerodynamic effects are entirely absent from these models, and so they lack fidelity in aggressive, post-stall maneuvering.

\begin{figure}[t]
\centering
\includegraphics[width=0.9\columnwidth]{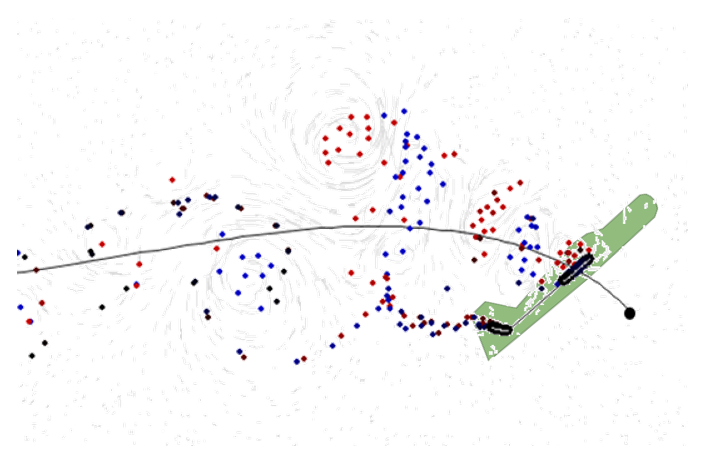}
\caption{A visualization of the discrete vortex simulator during a perching maneuver. The green UAV follows the black trajectory from left to right. The red and blue points are vortex particles representing the wake of the UAV---red for counterclockwise induced velocity, blue for clockwise. The total induced flow is indicated by the gray pathlines.}
\label{fig:vortex-shedding}
\end{figure}

In this work, we replace the quasi-steady aerodynamics in a UAV dynamics model with a vortex particle model, capable of representing unsteady aerodynamic effects (see visualization in Fig.~\ref{fig:vortex-shedding}) \cite{Katz2001}. Potential flow methods like the vortex particle model are dramatically more computationally efficient than grid-based computational fluid dynamics (at the expense of some physical fidelity), opening the possibility for simulating the fluid state in the control loop for real-time planning and state estimation \cite{Darakananda2018}. In this paper, we explore the performance improvements afforded by including these unsteady models in controller synthesis and execution. We evaluate our methods on aggressive, post-stall maneuvers as well as under dynamic wing-morphing. We show that including a computational model of unsteady aerodynamics to generate the control policy enables improved performance for a UAV executing a post-stall maneuver with and without wing morphing. While our control approach is not yet fast-enough for real-time re-planning, we believe that future computational advancements could allow receding-horizon methods that leverage simulation-in-the-loop. 

In summary, the main contributions are this paper are
\begin{itemize}
\item A novel discrete vortex simulator capable of running faster than real-time for planarized maneuvers.
\item A control algorithm capable of leveraging the discrete vortex model for trajectory optimization and feedback.
\item Demonstration of improved perching performance in hardware on a dynamic morphing-wing vehicle.
\end{itemize}

Our paper is organized as follows: Section~\ref{sec:UAV} introduces the morphing-wing UAV, Section~\ref{sec:aero} reviews the aerodynamic model, Section~\ref{sec:planning} describes its use in model-based control, and Section~\ref{sec:perch} presents the experimental results of UAV performance in gliding perch maneuvers.

\section{Related work}

Early work in post-stall maneuvers for autonomous UAVs aimed to hover a fixed-wing UAV by transitioning to and from a prop hang \cite{Sobolic2009}.
This was accomplished through linear feedback designed using a nonlinear UAV dynamics model which was improved by wind tunnel measurements of the test vehicle.
Further development of hybrid theoretical and empirical dynamics models was done in \cite{Khan2016}.
Their dynamics model was purpose-built for aerobatic fixed-wing UAVs performing maneuvers at high incidence angles and was used in related works to control agile UAVs in various aggressive maneuvers \cite{Levin2017, Bulka2018}.
Even these state-of-the-art models are unable to predict UAV dynamics with fidelity sufficient for long horizon planning.
In some cases, agile maneuvers with fixed-wing UAVs are accomplished by improving controller robustness rather than (or in addition to) model fidelity. For instance, in \cite{Moore2014}, the authors use the Linear Quadratic Regulator (LQR)-Trees algorithm to generate a library of feedback policies to improve robustness to initial conditions and model error. In \cite{Basescu2020}, model error is compensated for by an inner loop of Time Varying LQR (TVLQR) feedback and an outer loop of Nonlinear Model Predictive Control (NMPC) re-planning.

Recent work at the limits of autonomous quadcopter maneuverability has emphasized a theoretical hurdle which is even more apparent for agile fixed-wing UAVs: quasi-steady aerodynamics models are insufficient for high fidelity flight dynamics of extremely agile UAVs \cite{Bauersfeld2021}. Their improved dynamics model was produced by augmenting the quasi-steady physics-based model with a learned residual conditioned on the recent history of vehicle states.

There is yet little overlap between the aerodynamic models used by roboticists in control of agile UAVs and the modeling tools developed by aerodynamicists with agile UAVs as motivation.
Of the latter, recent work in potential flow models motivates the current work in applying computational aerodynamics to UAV control.
A review of various potential flow models designed for unsteady aerodynamics around UAVs can be found in \cite{Manar2019}.
One such model was developed in \cite{Ramesh2014}, augmented with state estimation via pressure sensors in \cite{Darakananda2018} and \cite{Provost2021}, and tested in predicting wing--gust response in \cite{Eldredge2021}.
Improving the physical fidelity, computational efficiency, and state estimation methods for potential flow models remain areas of active research \cite{Manar2019, Eldredge2021}.

\section{Morphing-wing UAV} \label{sec:UAV}

\begin{figure}[t]
\centering
\begin{subfigure}[b]{0.82\columnwidth}
\centering
\includegraphics[width=\textwidth]{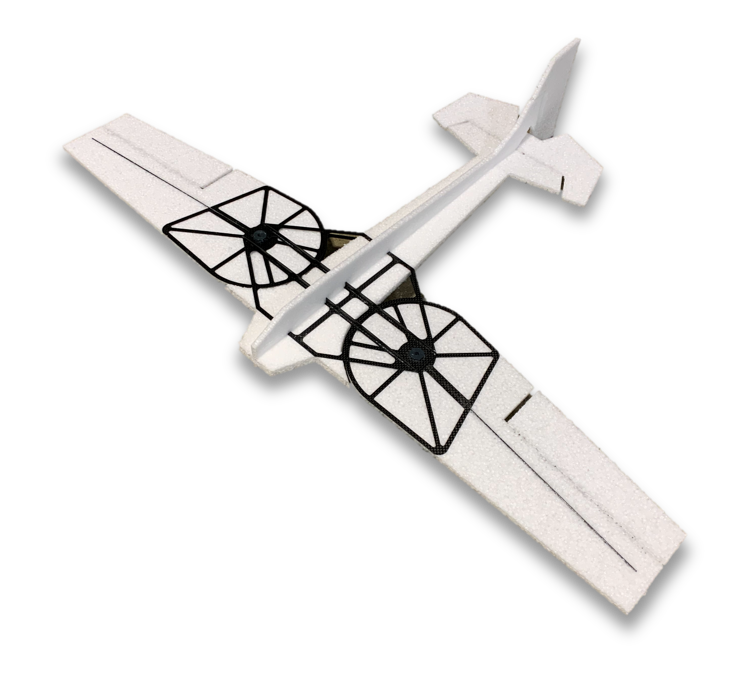}
\caption{Wings held in neutral position.}
\label{fig:UAV-a}
\end{subfigure}
\begin{subfigure}[b]{0.7\columnwidth}
\centering
\includegraphics[width=\textwidth]{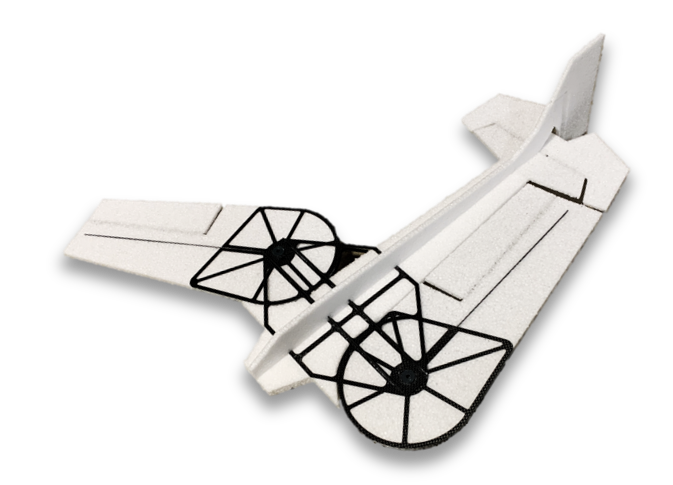}
\caption{Wings actuated to fore and aft limits.}
\label{fig:UAV-b}
\end{subfigure}
\caption{The morphing-wing UAV used in the present work.}
\label{fig:UAV}
\end{figure}

Aerobatic fixed-wing UAVs are capable of post-stall maneuvering, leveraging large control surfaces and significant thrust-to-weight ratio to maintain control authority.
However, fixed placement of aerodynamic surfaces relative to center of mass limits the agility of even these platforms.
In the current work, we explore the increased flight performance of a morphing-wing UAV.
While many forms of wing morphing are possible (and many are potentially beneficial), we focus on dynamic wing sweep.
Actuating wing positions significantly affects roll and pitch by moving the center of pressure, but is generally less mechanically complex than wing morphing by changing size or shape of lifting surfaces.

Fig.~\ref{fig:UAV} shows our morphing-wing UAV.
It is a small, carbon-reinforced foam aircraft designed around a conventional planform.
It has a \SI{70}{\centi\meter} wingspan and \SI{200}{\gram} mass (including the electronics not visible in Fig.~\ref{fig:UAV}).
In addition to conventional control surfaces---rudder, elevator, and two ailerons---the wing sweep angles are controlled by servos near the root of each wing.
The propeller was not included for any of the experiments in this work.

The two shoulder degrees of freedom of this morphing-wing UAV increase the potential agility of the platform at the expense of dynamic stability, motivating our need for greater aerodynamic model fidelity.
In their un-actuated position, the wings' leading edges are orthogonal to the fuselage (as they are in Fig.~\ref{fig:UAV-a}).
Relative to that position, each wing can swing independently up to \SI{30}{\degree} toward the UAV's nose, and up to \SI{90}{\degree} toward the tail.
Traversing the full range of motion takes \SI{0.2}{\second}.
The morphing wings and the ailerons were not simultaneously controlled in our experiments; flights with active wing morphing held the ailerons fixed in their neutral position.

Autonomous control is computed off-board on a laptop computer and radioed to the UAV.
Dynamics simulations for sample-based control (see sections \ref{sec:aero}, \ref{sec:planning}) are computed in efficient, vectorized operations in the NumPy Python library \cite{Harris2020}.
Optimized controls are communicated through C++ Robot Operating System (ROS) modules \cite{Quigley2009}.
Vehicle state is provided to the controller by an OptiTrack motion capture system.
The conventional control surfaces are actuated by BMS-101HV servos, and the wing morphing by KST X08 servos.
The UAV is powered by a Crack 2S 180 mAH LiPo battery.

\section{Unsteady aerodynamics} \label{sec:aero}
Post-stall aerodynamics pose a problem for quasi-steady models, since the local aerodynamics vary over time even for fixed vehicle attitude.
These variations are not randomness in the dynamics; they reflect a dependence of the forces and moments on conditions not represented by (or deducible from) the vehicle attitude.
A flight dynamics model without aerodynamic state represented can only be accurate on average, and may still deviate significantly at any particular moment.
For low inertia UAVs, neglecting the time-varying dynamics can significantly limit overall model fidelity.
The goal of this work is to improve flight dynamics fidelity in post-stall conditions by explicitly modeling the local aerodynamics.

Using any of the many techniques available for Computational Fluid Dynamics (CFD), the aerodynamics around the flying UAV can be computed to arbitrary precision by numerically solving the discretized Navier-Stokes equations.
However, these methods are not remotely intended for real-time model-predictive control, and typically take hours or days of computation for seconds of simulation.
Instead, computationally-efficient modeling of unsteady aerodynamics relies on data-driven solutions or on solving simplified governing equations.
Potential Flow models simplify the Navier-Stokes equations such that flows can be expressed as sums of solutions to LaPlace's equation: $\nabla^2\phi=0$ where $\nabla\phi=\mathbf{v}$; $\phi$ is the scalar field of flow potential, and $\mathbf{v}$ is the vector field of flow velocity.
Numerical potential flow models---such as the vortex particle method used in this work---take advantage of this to represent flows using a Lagrangian discretization, a collection of fluid elements, rather than discretizing the domain using a Cartesian grid.
This representation requires far less computation than grid-based methods, and so provides a promising avenue for real-time control with unsteady aerodynamics.

\subsection{Vortex Particle model}
In this work, the aerodynamic state near the UAV is represented using vortex particles, Lagrangian elements of flow carrying vorticity.
Each vortex particle induces flow velocity in concentric circles around itself and is moved by the local flow velocity.
The sum of velocity induced by all vortices is the flow solution modeled by the vortex particle method.
In our model, the influence of the wing is \emph{also} represented by a collection of vortex particles.
The wing's vortices are ``bound'' to the surface; they do not move with the flow as the wake vortex particles do.

\begin{figure}[tb]
\centering
\includegraphics[width=0.95\columnwidth]{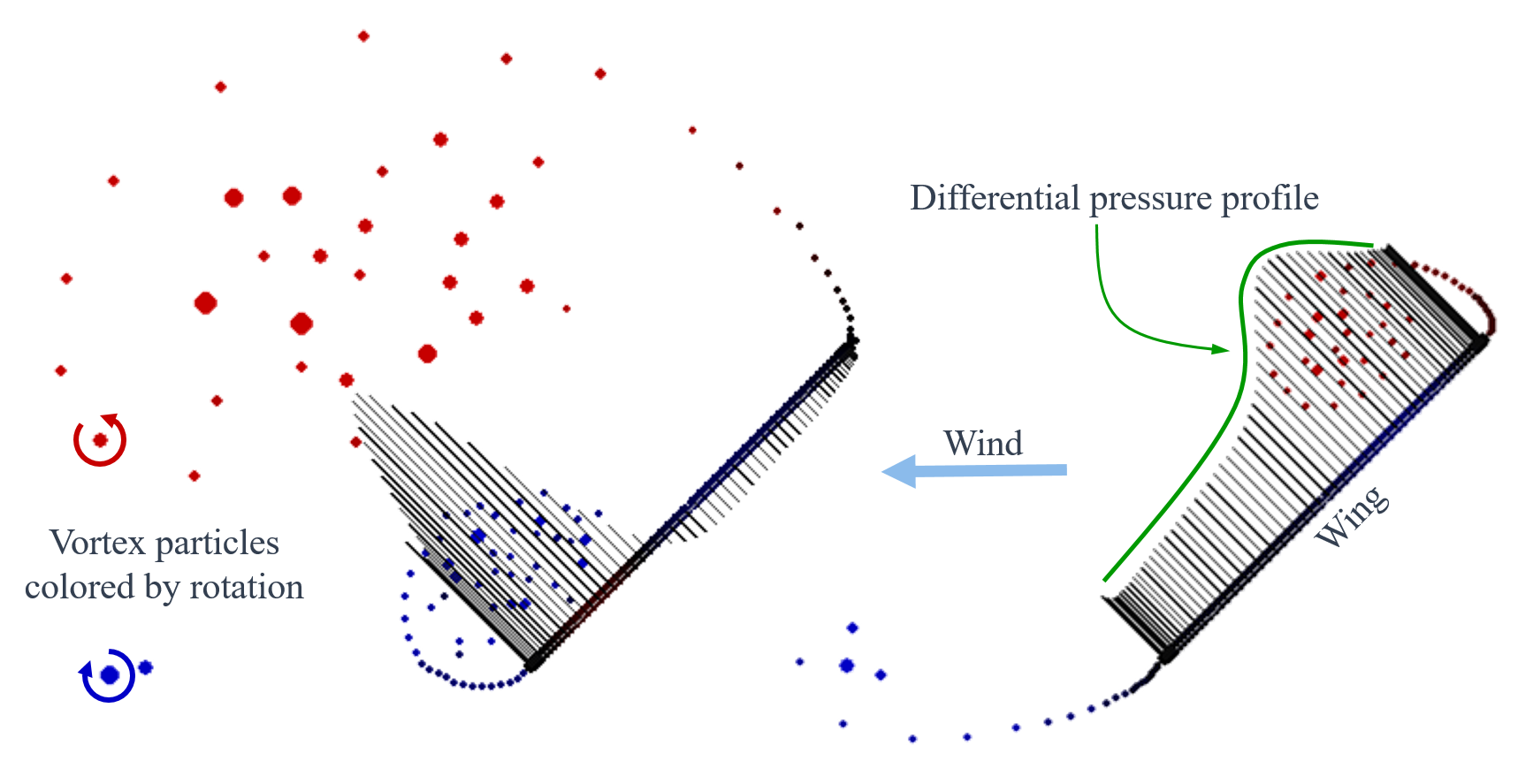}
\caption{Two examples of the vortex particle model representation of local aerodynamics for thin wings. The thick black line is the wing, and the red and blue points are vortex particles---red for counterclockwise induced velocity, blue for clockwise. In both cases, the wing is held at \SI{45}{\degree} incidence to oncoming, constant wind. Despite identical pose and velocity, the two cases experience significantly different forces due to the time-varying local aerodynamics.}
\label{fig:DVM}
\end{figure}

Fig.~\ref{fig:DVM} shows the two-dimensional aerodynamic state around a thin, flat wing at \SI{45}{\degree} incidence to the oncoming wind.
The boundary condition imposed by the wing causes shear against the surface which results in local vorticity in the air.
At each simulated time step, the vorticity released from the wing section into the air is captured as one new vortex particle at the leading edge, and another at the trailing edge.
(Leading edge vortex shedding for airfoils can be quite complex, but in cases of thin, sharp edges, it can be assumed that vorticity is continuously produced \cite{Sheng2005}.)
The existing particles convect with the local velocity, and in this case they roll up into larger coherent structures.

Each vortex has a scalar strength, $\Gamma$, and a two dimensional position, $\mathbf{x}$.
The velocity induced by a vortex particle on a target location is

\begin{equation}
\label{eq:kernel}
\mathbf{v}_\mathrm{target} =
\frac{\Gamma_\mathrm{vortex}}{2\pi r^2}
\left[\begin{array}{cc}
0&1\\
-1&0\\
\end{array}\right]
\Delta\mathbf{x}
\end{equation}

where $r = ||\Delta\mathbf{x}||$ and $\Delta\mathbf{x} = \mathbf{x}_\mathrm{target} - \mathbf{x}_\mathrm{vortex}$.
In our model, the wake vortices (but not the bound vortices) use a modified form of this influence kernel, which represents the finite core of a real vortex in viscous fluid.
This is accomplished by scaling (\ref{eq:kernel}) such that
\begin{equation}
\label{eq:kernel2}
\mathbf{v}_\mathrm{target} =
\frac{\Gamma_\mathrm{vortex}}{2\pi r^2}
\frac{\left( \frac{r}{r_\mathrm{core}} \right)^2 }
{\sqrt{ 1 + \left( \frac{r}{r_\mathrm{core}} \right)^4 }}
\left[\begin{array}{cc}
0&1\\
-1&0\\
\end{array}\right]
\Delta\mathbf{x}
\end{equation}
where $r_\mathrm{core}$ is the specified core radius of the model \cite{Leishman2006}.

\newcommand{\Nbound}{n_\textrm{bound}}

The wing's surface is defined by $\Nbound$ bound vorticies and $\Nbound + 1$ control points at which the surface boundary condition is enforced.
In this model, $\Nbound - 1$ control points are centered between each pair of adjacent bound vortices and the remaining 2 are located at the leading and trailing edges of the wing.
The boundary condition, ``no through flow,'' requires that there be zero relative velocity of fluid and surface normal to the surface.

\begin{equation}
\Delta\mathbf{v} \cdot \hat{\mathbf{u}}_\mathrm{normal} = 0
\end{equation}
where $\Delta\mathbf{v} = \mathbf{v}_\mathrm{surface} - \mathbf{v}_\mathrm{air}$ and $\hat{\mathbf{u}}_\mathrm{normal}$ is the unit vector normal to the surface.

The scalar strength value of each bound vortex is solved to satisfy this condition at each control point.
The surface edges also shed vorticity into the fluid, creating one new wake vortex particle each at every time step.
Together, the strengths of the bound vortices and new wake vortices make $\Nbound + 2$ free variables.
The final criterion which closes the system of equations is Kelvin's theorem, which states that total circulation is conserved.

\begin{equation}
\frac{d}{dt}\Gamma_\mathrm{total} \equiv \frac{d}{dt}\sum \Gamma_i = 0
\end{equation}

where $\Gamma_\mathrm{total}$ is the system's total circulation and $\Gamma_i$ is the strength of a particular vortex particle.

Every vortex influences the velocity at each control point proportional to its strength, $\Gamma$, so the boundary conditions can be solved as a system of $\Nbound + 2$ linear equations.
This solution fixes the strengths of the new wake vortices and specifies the current strength of each bound vortex.

The UAV dynamics are determined from the local aerodynamics by computing the pressure difference across the surface.
Following \cite{Katz2001}, the pressure difference at the $i$th bound vortex from the leading edge is
\begin{equation}
\Delta p_i = \rho
\left[
\Delta\mathbf{v}_i
\cdot \hat{\mathbf{u}}_{i, \mathrm{tangent}}
\frac{\Gamma_i}{\Delta l_i}
+ \frac{d}{dt}\sum_{j=0}^{i}\Gamma_j
\right]
\end{equation}
where $\rho$ is the fluid density, $\hat{\mathbf{u}}_{i, \mathrm{tangent}}$ is the surface-tangent unit vector at $i$, and $\Gamma_i$ and $\Delta l_i$ are the total vortex strength and length along the surface associated with the $i$th bound vortex.
The term $\frac{d}{dt}\sum_{j=0}^{i}\Gamma_j$ represents the path integral of circulation rate-of-change from ambient flow to the $i$th bound vortex along the wake and surface, including the newest leading edge wake vortex at $i=0$.
Integrating this pressure distribution across the surface produces the forces and moments on the UAV, which in turn are used to update the vehicle's velocity and pose for the next time step.
The position of each wake vortex is simultaneously updated based on the local fluid velocity.

Over many time steps of this process the number of wake vortices grows to an unmanageable quantity.
Computational efficiency is maintained by identifying pairs of vortices which can be merged together with minimal error introduced.
Specifically, we replace pairs of vortices with one vortex when doing so changes the induced velocity on the wing less than a threshold amount.
The new vortex has strength equal to the sum of the replaced vortices' strengths, and is located at their strength-weighted-average position.

The model discussed so far represents two dimensional unsteady aerodynamics around thin surfaces.
To model the whole UAV, we used multiple parallel slices of the two dimensional representation and added quasi-steady contributions for non-lifting surfaces.
This ``two-and-a-half'' dimensional model is limited to small out of plane velocity; generalization to large side-slip would require further model development.

\section{Planning and Control}
\label{sec:planning}
\begin{figure}[tb]
\centering
\includegraphics[width=0.9\columnwidth]{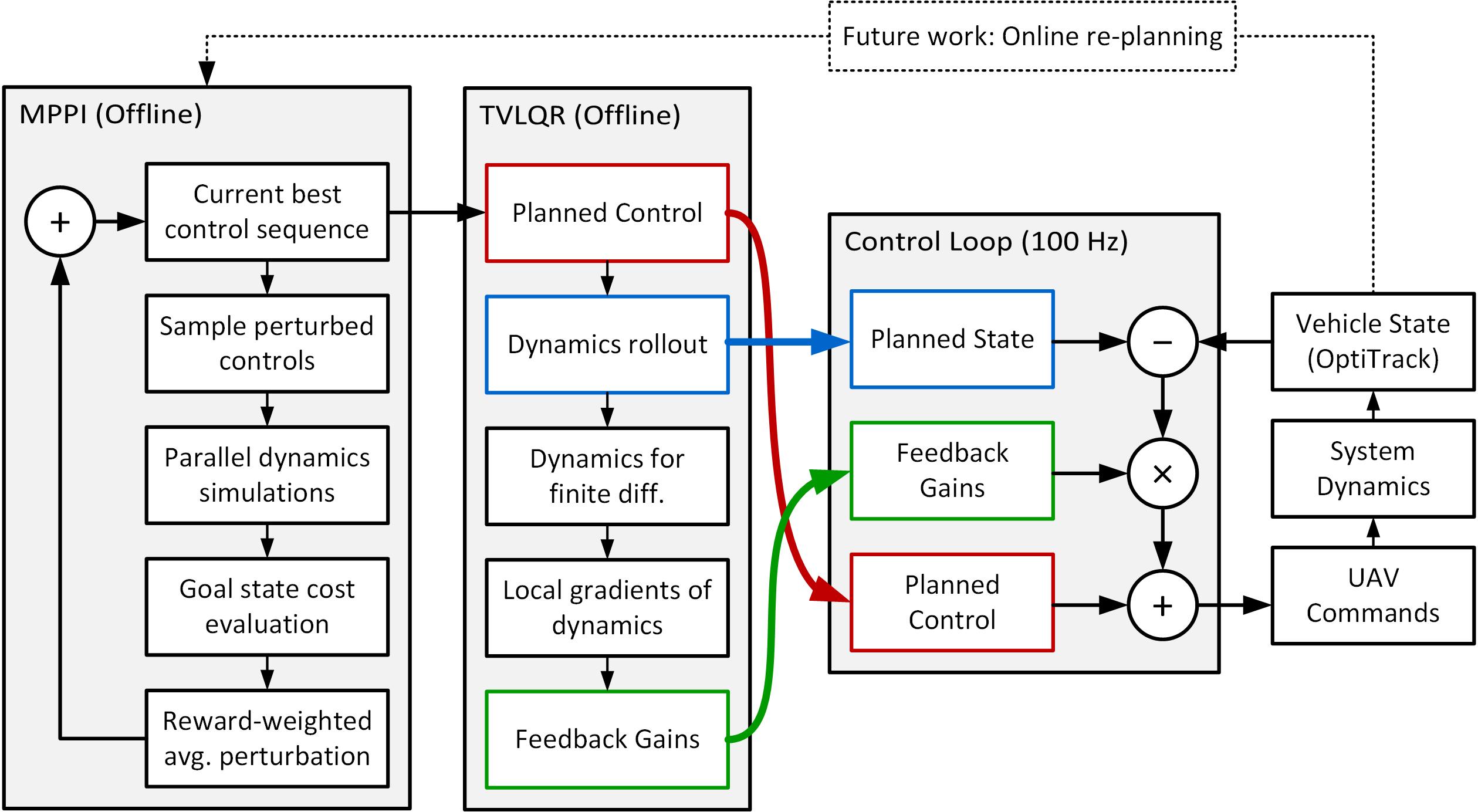}
\caption{Flowchart of our planning and control algorithm.}
\label{fig:Flowchart}
\end{figure}

We envision this vortex particle model used as part of a receding-horizon NMPC algorithm comprised of an outer loop of trajectory optimization and an inner loop of locally-linear feedback similar to \cite{Basescu2020}.
This algorithm is summarized in figure \ref{fig:Flowchart} and described in detail in this section.
Due to the computation time requirements of the vortex particle model, our current work only explores a single execution of that outer loop, which permits pre-planning of the maneuver.
This reduces the problem to off-line trajectory optimization. However, with additional model development, these results could generalize to a full receding-horizon NMPC implementation.

\subsection{Model-predictive control}
The vortex particle model augments the UAV's state space with a variable-length, arbitrarily-large characterization of local aerodynamics.
This is not easily compatible with direct optimization techniques for maneuver planning which include the state variables as decision parameters.
Instead, the perch maneuver was planned using a sampling-based approach which only uses inputs rather than states as decision parameters.
We adapt the Model Predictive Path Integral (MPPI) approach of \cite{Williams2017}, which relaxed restrictions on the dynamics model compared to previous MPPI approaches.

We specify initial and target conditions for the UAV, $X_0$ and $X_\textrm{target}$, and initial conditions for the vortex particle model and control sequence (both of which begin with all zeros in this work's launch--glide--perch experiment presented in section~\ref{sec:perch}).
At each iteration, many perturbations to the nominal control sequence are generated by sampling from zero-mean normal distributions.
The flight dynamics are simulated using the vortex particle model to observe the state trajectory resulting from each input sequence.
Each trajectory is assigned a cost, which here is simply a scaled quadratic cost computed from the closest approach to the target.

\begin{equation}
\label{eq:cost}
\textrm{cost} = \min_\textrm{over traj.} \left(
\left|\left|\begin{array}{c}
\Delta x \\ \Delta z \\ \Delta \alpha
\end{array}\right|\right|^2
+ 0.2 \left|\left|\begin{array}{c}
\Delta \dot{x} \\ \Delta \dot{z} \\ \Delta \dot{\alpha}
\end{array}\right|\right|^2 
\right).
\end{equation}

\begin{figure}[tb]
\centering
\includegraphics[width=0.95\columnwidth]{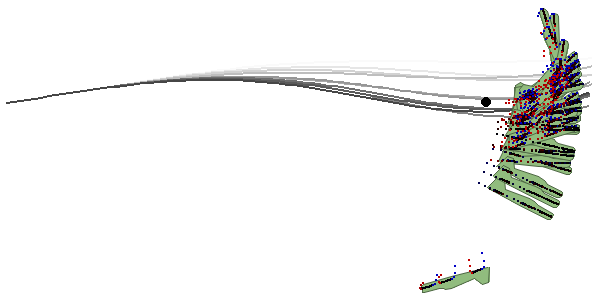}
\caption{Render of the dynamics roll-outs for perturbed control sequences used for sample based planning. The green airplane silhouettes reflect the UAV state of each sampled trajectory, and the surrounding red and blue points are the vortex particles. The gray lines show the expected trajectory of the nominal control sequence which improves gradually over each iteration. Trajectories begin on the left and aim to reach the virtual perch target, displayed as a black circle, on the right. Darker trajectory lines represent more recent nominal paths.}
\label{fig:mppi}
\end{figure}

During path planning, the out-of-plane state components, $y$, $\phi$, and $\psi$, are constrained to zero by model symmetry.
The nominal control sequence is replaced by an exponential weighted average of the perturbed controls, where the weights are negative cost.
A render of the converging nominal trajectory and simulated states for perturbed controls is shown in Fig.~\ref{fig:mppi}.

The combination of vortex particle model and sampling-based planner is computationally demanding.
Our model is implemented such that all particle interactions ((\ref{eq:kernel}) and (\ref{eq:kernel2})) and all parallel samples for MPPI are computed as a single vectorized operation, optionally executed on a graphics card.
Additionally, there are many adjustments in the model which balance performance and computational expense.
Physical fidelity of the vortex particle method requires adequate temporal and spatial resolution and can be assisted by multi-step numerical integration.
Convergence of the control sequence depends on an adequate number of sampled perturbations per iteration and an adequate number of total iterations.
Reducing any of these parameters increases the computational speed, but could compromise model fidelity or control convergence.
We were unable to find a test configuration where the combined model and planner produce meaningful control sequences in real time for receding horizon planning.
Instead, maneuvers for the presented experiments had to be planned ahead of flight time using known initial conditions.
On the laptop used for those experiments, planning a \SI{1.5}{\second} maneuver took at least \SI{5}{\second} of computation.
(Keep in mind that the start of the control sequence is executed almost immediately, so the planner must return in far less time than the planning horizon.)
Once real-time control was ruled out, the model resolution and planner samples and iterations were increased for the experiments discussed in Section~\ref{sec:perch}; the final maneuver was planned in \SI{1}{\minute}.

\subsection{Feedback control}
To generate our local linear feedback policy for trajectory tracking, we employ TVLQR. Because we cannot directly compute a gradient through the vortex particle model, we simulate the nominal state trajectory and the local partial derivative of dynamics with respect to each UAV state and control variable through finite difference at intervals along that trajectory.
This procedure requires careful managing of the vortex particle model state, but is otherwise identical to TVLQR for quasi-steady aerodynamics models.
These partial derivatives contribute to a backwards integral over the trajectory which produces linear feedback gains for each point computed (see \cite{Basescu2020} for details on our implementation).
In flight, the executed commands are the combination of the nominal values from MPPI and the linear feedback from TVLQR.
In Fig.~\ref{fig:perchImg}, the morphing wings are actuated asymmetrically for a nominally-planar maneuver; this is the TVLQR response to our intentionally-perturbed initial conditions.

\section{Perch maneuver experiments} \label{sec:perch}
\begin{figure}[tb]
\centering
\includegraphics[width=\columnwidth]{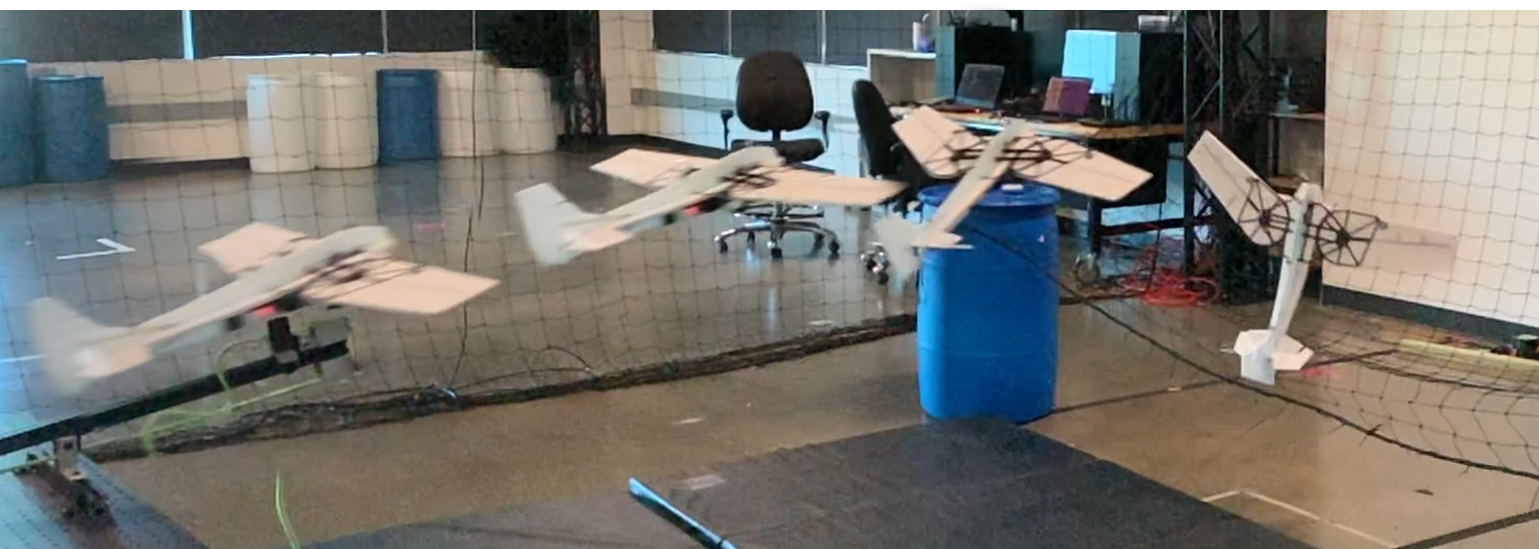}
\caption{Superimposed snapshots of the UAV in flight. The maneuver begins with launch from the elastic band on the left and ends at the virtual perch target just above the net on the right.}
\label{fig:perchImg}
\end{figure}

We experimentally tested the impact on autonomous flight performance of the morphing wings and the vortex particle model by comparing achieved target distance for a particular post-stall maneuver, launch--glide--perch.
Perching maneuvers require maintaining control while intentionally stalling a UAV's wings, and are a common choice of test case for experiments in nonlinear control \cite{Moore2014, Wickenheiser2008, Moore2012, Manchester2017}.
In our experiment, the UAV is launched using a guide rail and elastic cable for repeatable initial conditions.
Once the UAV clears the launcher, it has \SI{3.5}{\meter} to arrest its initial velocity of \SI{7}{\meter\per\second} and arrive at the perch target.
In addition to specifying the perch location, the controller aims to arrive at the perch with precise orientation and velocity: \SI{45}{\degree} pitch, no roll or yaw, \SI{0.5}{\meter\per\second} velocity forward and downward, and no velocity sideways.
There is no real perch mechanism at the target location; after the maneuver, the UAV falls into the arena net.
This UAV was flown in a motion capture arena using autonomous, off-board control.
Fig.~\ref{fig:perchImg} shows the UAV performing this maneuver in our motion capture facility.

\subsection{Performance results}
The same launch--glide--perch maneuver was repeated for four configurations of the UAV: (1) fixed-wing sweep with conventional aerodynamic model, (2) fixed-wing sweep with vortex particle model, (3) morphing wing with conventional aerodynamic model, and (4) the primary case of morphing wing with vortex particle model.
This permits investigation of the individual contributions to flight performance of the hardware and software modifications.
The perch maneuver was repeated 10 times for each of these four configurations.

\begin{figure}[tb]
\centering
\includegraphics[width=\columnwidth]{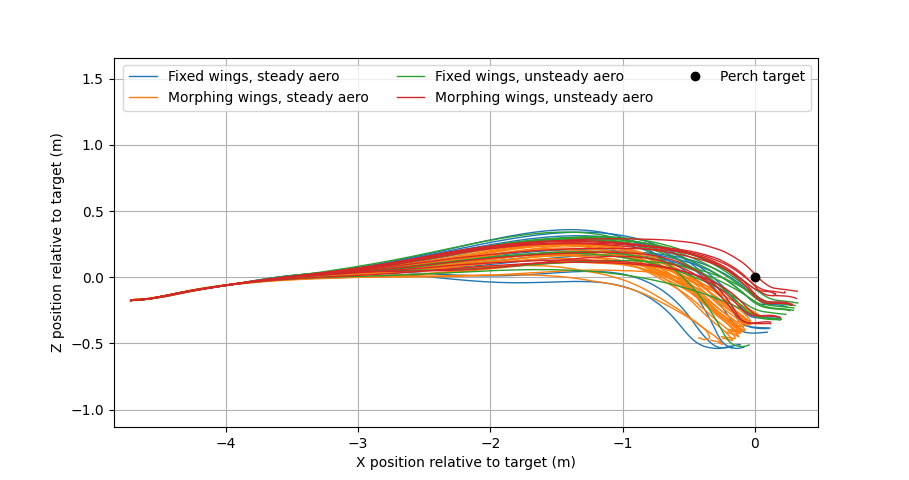}
\caption{Comparison of launch--glide--perch maneuver trajectories for four cases: fixed-wings and steady aerodynamics (the conventional case), morphing wings and steady aerodynamics, fixed-wings and unsteady aerodynamics (the vortex particle model), and morphing wings and unsteady aerodynamics. On this plot, the UAV flight is left to right.}
\label{fig:perchXY}
\end{figure}

Fig.~\ref{fig:perchXY} shows the trajectory of the UAV during each of the tested perch maneuvers.
Each line is colored by the configuration used for that sample.
Qualitatively, all configurations seem able to slow down quickly and fall near the perch target with varying precision.
For a quantitative comparison of perch maneuver performance, we compute the minimum target cost achieved on each test.
The quadratic cost function used in maneuver planning, (\ref{eq:cost}), is also used for evaluation.

The resulting minimum costs are shown in Table~\ref{tab:results}.
The absolute value of the cost is not meaningful, so each value is normalized by the mean of the baseline case.
Using the unsteady aerodynamic model without wing morphing did improve over the baseline performance slightly, and using both wing morphing and the unsteady model improved performance further.
However, the maneuver performance was far worse when attempting to plan for the morphing wing using the quasi-steady aerodynamic model.

\begin{table}[th]
\centering
\caption{Perch maneuver performance relative to baseline.}
\label{tab:results}
\footnotesize
\begin{tabular}{ccc}
Wing configuration & Aerodynamic model & Normalized cost \\
& & (mean $\pm$ std. dev.) \\ \hline
Fixed & Quasi-steady & $1.0 \pm 0.83$ \\
Fixed & Unsteady & $0.83 \pm 0.56$ \\
Morphing & Quasi-steady & $3.7 \pm 0.61$ \\
Morphing & Unsteady & $0.72 \pm 0.16$ \\
\end{tabular}
\end{table}

\section{Conclusion}
In this paper, model-based control of agile UAVs was extended to a morphing-wing UAV and an explicit representation of local aerodynamics.
Experimental comparison of performance in perching maneuvers suggests that the wing morphing degrees of freedom add to the UAV's maneuverability and that improvement of autonomous performance was enabled by the vortex particle model, an unsteady, first-principles-based computational aerodynamics model.
The wing morphing capability did not improve performance in paths planned using a conventional, quasi-steady aerodynamics model.

Our results support the notions that greater agility is achievable from fixed-wing or morphing-wing UAVs and that autonomous realization of this capability benefits from a dynamics model conditioned on unsteady aerodynamics. However, our vortex particle model is not capable of real-time planning given current computational resources. Future work will explore modifications to our model formulation to improve computational speed with the goal of enabling real-time planning. This would face a remaining challenge in state estimation for the vortex particle model as discussed in \cite{Darakananda2018}, so another future research direction is the exploration of pressure sensing on the UAV wings to increase vortex particle state observability. Finally, we plan to investigate data-driven approaches for modeling unsteady effects. As explored in \cite{Bauersfeld2021}, machine learning may provide an alternate means of producing high fidelity UAV dynamics models conditioned implicitly on the local aerodynamics.

\bibliographystyle{IEEEtran}
\bibliography{IEEEabrv, morphingwing}

\end{document}